\documentclass[times, 10pt,twocolumn]{article}
\usepackage[utf8]{inputenc}
\usepackage{latex12}
\usepackage{times}
\usepackage{balance}
\usepackage{amsmath}
\usepackage{amssymb}
\usepackage{bm}
\usepackage{subfig}
\usepackage{graphicx}
\usepackage{url}
\usepackage{booktabs}
\usepackage{needspace}
\usepackage{microtype}

\begin{document}

\title{Local Water Diffusion Phenomenon Clustering From High Angular
Resolution Diffusion Imaging (HARDI)\thanks{This work is supported by Samuel de Champlain 63.102 funds}}

\author{Romain Giot and Christophe Charrier\\
\emph{Universit\'e de Caen}, 
\emph{ENSICAEN}, 
\emph{CNRS}\\
\emph{UMR 6072 GREYC}\\
\emph{romain.giot@ensicaen.fr}\\
\emph{christophe.charrier@unicaen.fr}
%
\and{Maxime Descoteaux}\\
\emph{Sherbrooke Connectivity Imaging Laboratory}\\
\emph{Computer Science department}\\
\emph{Universit\'e de Sherbrooke}\\
\emph{m.descoteaux@usherbrooke.ca}
}

\maketitle
\thispagestyle{empty}

\begin{abstract}
The understanding of neurodegenerative diseases undoubtedly
passes through the study of human brain white matter fiber tracts. To date,
diffusion  magnetic resonance imaging (dMRI)  is the unique technique
to obtain information about the neural architecture of the human
brain, thus permitting the study of white matter connections and their
integrity.  However, a remaining challenge of the dMRI community is to
better characterize complex fiber crossing configurations, where diffusion
tensor imaging (DTI) is limited but high angular resolution diffusion
imaging (HARDI) now brings solutions. This paper 
investigates the development of both identification and
classification process of the local water diffusion phenomenon based on
HARDI data to automatically detect imaging voxels where there are
single and crossing fiber bundle populations. The technique is based on
knowledge extraction processes and is validated on a dMRI phantom
dataset with ground truth.
\end{abstract}

\vspace*{-0.5cm}
\section{Introduction} 
Diffusion Weighted magnetic resonance imaging (DW-MRI) is able to
quantify the anisotropic 
diffusion of water molecules in biological tissues such as the human
brain white matter.  The great success of DW-MRI comes from its
capability to accurately describe the geometry of the underlying
microstructure. 
DW-MRI captures the average diffusion of water
molecules, which probes the structure of the biological tissue at
scales much smaller than the imaging
resolution~\cite{descoteaux-poupon:12}.  
New dMRI techniques for high angular resolution
diffusion imaging (HARDI) are now able to
recover one or more directions of fiber populations at each imaging
voxel and thus, overcome some of the limitations of diffusion tensor
imaging (DTI) in regions of complex fiber configurations where fibers
cross,  branch and kiss. 

\begin{figure}
  \centering
  \includegraphics[width=.9\linewidth]{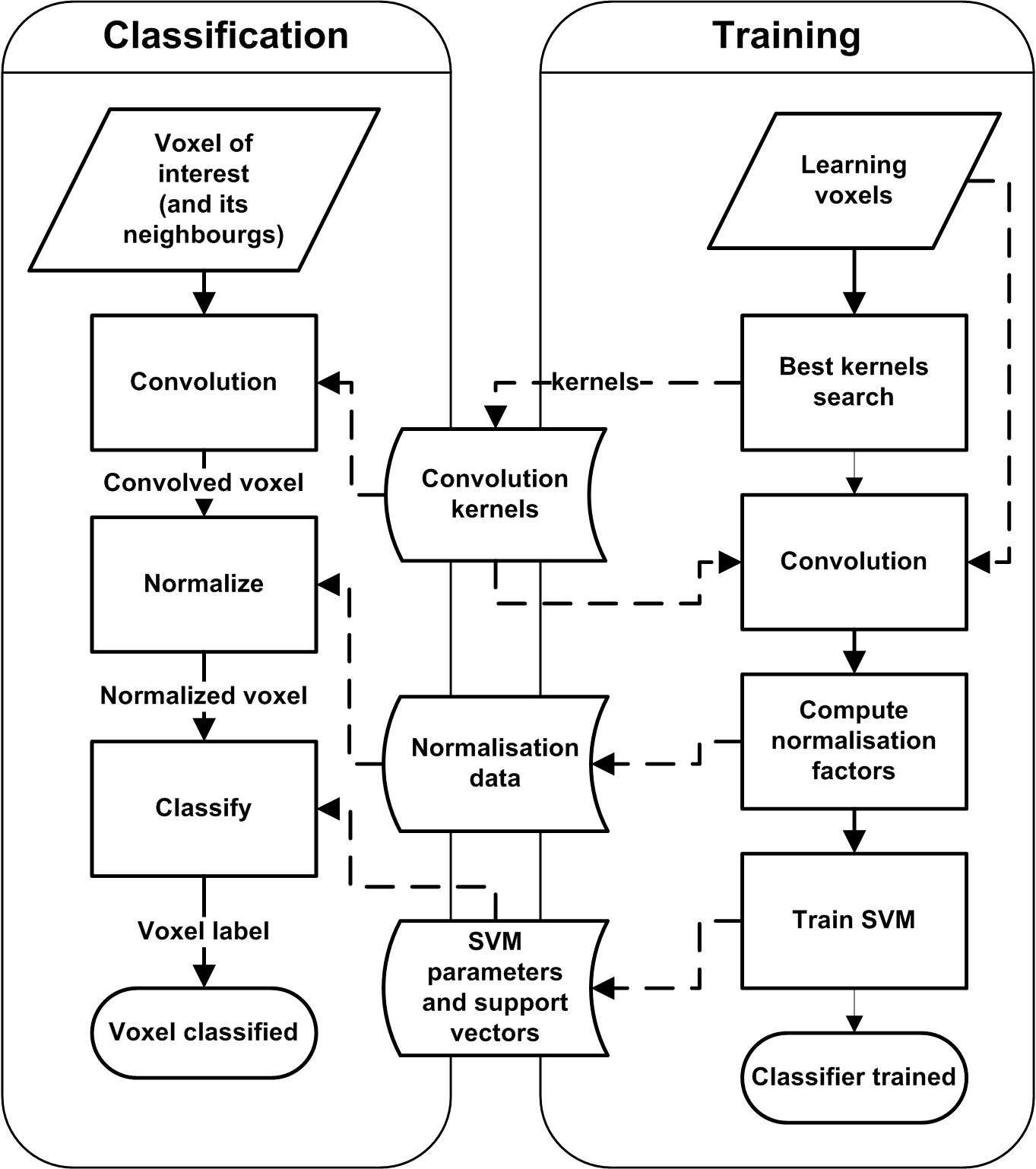}
   \caption{Proposal}
\end{figure}

Most current classification techniques are based on  DTI
measures of Westin et al~\cite{Westin2002}. For example, a recent 
paper from~\cite{Vos2012} 
classifies the white matter voxels into single and crossing fibers
simply by hard thresholding the linear, planar and spherical measures
computed from the eigen-values of the diffusion tensor.  Other
techniques have been developed to better handle fiber crossings using
apparent diffusion coefficient~\cite{descoteaux-angelino-etal:06c}
modeling from HARDI and other HARDI
models representations based on spherical harmonics
(SH) decomposition~\cite{prckovska09}.     
It was previously
  suggested using automatic Bayesian relevance
  determination~\cite{behrens-etal:07} that 
  between 1/3 and 2/3 of WM voxels contained crossings. Jeurissen et
  al~\cite{jeurissen-leemans-etal:12} have 
  just showed that this number is an underestimation and that
  WM crossings can take up to 90\% of WM
  voxels using maxima
  extracted from a robust SH-based 
  fiber orientation distribution estimation using constrained
  spherical deconvolution.
Otherwise, the first attempts to classify HARDI voxels using
machine learning techniques was done
in~\cite{wassermann-descoteaux-etal:08} using the space of SH
coefficients from q-ball imaging. These techniques are based on
diffusion maps and spectral clustering but were never applied to
neurodegenerative datasets. More recently, Schnell et al.~\cite{schnell-saur-etal:09}
have designed a classification process based on support vector
machines, also based on the SH representation of the dMRI 
signal and have compared their results with the classical
classification of Westin's measures, as done in~\cite{Vos2012}.

The main contribution of this paper is the creation of a system
allowing to automatically classify voxels of HARDI data in order
to segment the white matter among different classes.
The proposed method goes deeper than the approach presented
in~\cite{schnell-saur-etal:09} because it takes  
into account the neighbourhood of the voxels by using convolved
data.


\section{Classifier Creation and Evaluation}
\subsection{Proposed Approach}
Using the Spherical Harmonic (SH) representation of each
voxel~\cite{descoteaux-angelino-etal:06c}, we want to be able to 
classify it as: 
(a) white matter with a single fiber bundle (WMSF), 
(b) white matter with crossing fiber bundles (WMCF),
(c) non-white matter (N-WM) of type gray matter (GM) or
cerebrospinal fluid (CSF).
Section~\ref{sec_apparatus} presents the dataset.
We use a Support Vector Machine (SVM)~\cite{vapnik1996theory} classifier to obtain the label of a voxel.
The SVM tries to find the best separating hyperplane between two
classes. As we work with several classes, a one-against-one approach is
used (we train $n\_class*(n\_class-1)/2$ classifiers).

We easily see that using only the SH information may be not enough accurate, as
there is no knowledge of the neighborhood while classifying the voxel.
To address this problem, we operate a 2D convolution of each slice of the brain
against a kernel.
This allows to work with a voxel representation resulting of a weighted sum of the neighboring
voxels.
The convolution kernels are chosen in order to obtain the best accuracy (see
Section~\ref{sec_selection}).

\subsection{Classification of Voxel Using a Particular Feature
Space and its Kernels}
Say we have 
$\mathcal{F}$ the selected
feature space of cardinality $n$ (\emph{i.e.}, each voxel is represented by a
vector of size $n$ ; $n$ depends on the SH order) and
$\mathcal{K} = \{k_1, \cdots, k_n\}$, the convolution kernels with
one convolution kernel per dimension of the features space.
We apply the convolution kernel $k_i$ on each feature $i$ of each voxel of
each slice $j$:
\begin{equation}
vc_i^j = v_i^j \otimes k_i
, \forall i \in [1,n], \forall j \in [1,N]
\end{equation}
with $v_i^j[x,y]$ the feature $i$ of the voxel located at position
$(x,y)$ of the slice $j$,
$vc^j$ the slice $j$ after convolution of all voxels and
features and $N$ the number of slices.
Now, we consider all the voxels as being in a whole set $\Omega$ of couples
$\{\mathbf{x}, l\}$ ($\mathbf{x}$ is the feature vector and $l$ its
label) whatever their
slice and localisation in the slice:
\begin{equation}
\Omega=\bigcup_{j,x,y}\{vc^j[x,y],
label^j[x,y]\}
\end{equation}
with $label^j[x,y]$ the label of the voxel at ($x,y$) in slice
$j$).

We apply a 6 fold stratified cross validation (we obtain 6 subsets, where each
label is represented at the same ratio in each subset: there is the same
recognition difficulty in each subset).
Each subset serves as a testing set, while the other subsets serve as training set
(the following procedure is then applied 6 times).
The mean ($\bm \mu$) and the standard deviation ($\bm \sigma$) of
the feature vectors of the training set is
computed in order to apply a zscore normalisation of the training and testing
sets ($\mathbf{normalised\_sample} = (\mathbf{sample} - \bm \mu) / \bm \sigma$).
The normalised training samples serve to train a SVM with a gaussian kernel.
To quickly evaluate the couple ($\mathcal{K}$, $\mathcal{F}$) we do not try to
search the best SVM parameters. So $\gamma$ and $C$ are at the default value of
libsvm ($C=1, \gamma=1/n$).
The learned SVM is used to predict the label of the testing samples.

\subsection{Evaluation of a Set Kernels for a Feature Space}
There is a large disparity in the number of individuals of each class. 
So, we do not want to simply compute an averaged classification
error rate.
Instead we compute the following errors: 
(i) \emph{Missed WM Ratio (MWMR)}: ratio of WMSF and WMCF voxels
recognized as being N-WM (CSF or GM) voxels;
(ii) \emph{Exchanged WM Ratio (EWMR)}: ratio of WMSF voxels recognized as
being WMCF voxels and WMCF voxels recognized as WMSF voxels;
(iii) \emph{Imagined WM ratio (IWMR)}: ratio of N-WM (CSF
and GM) voxels recognised
as being WMSF or WMCF voxels.
Then the final error score is computed as following: 
\begin{equation}
  \label{eq_fitness}
rate=\alpha*MWMR + \beta*EWMR + \gamma*IWMR
\end{equation}
\noindent with $\alpha=1.5$, $\beta=1$ 	and $\gamma=2$ in this
experiment.
This way, we give a low weight on the EWMR and bigger weights
on IWMR and MWMR.
Although we try to differenciate them, we do not care of exchanging CSF and GM.

\subsection{Selection of the Best Kernels}
\label{sec_selection}
As, for each feature set, the search space is very huge, we use a
genetic algorithm in order to search the 
best convolution kernels.
Each kernel is an array of size $w*w$, with $w$ the width of the
kernel.
The genome consists of the $n$ kernels stored in a 1D array using
values in the interval $[-2;2]$ (it means that values inside the
kernel will be in this interval), so it has a size of $w*w*n$.
We do not pay attention to have a kernel summing to $1$ as we do
not care if the data obtained after convolution is not in the same
metric as the original data (we do not manipulate or visualize
the convolved data, it only serves for classification).
Population contains 500 individuals.
The initial population contains:
\begin{itemize}
	\item  one individual build with gaussian kernels which
	are supposed to be good by doing a weighted average based on the distance around
the voxel of interested
        \item 250 individuals created with modification of the
first individual (mean of the gaussian kernels based individual and a random one
      \item  249 individuals generated totally randomly.
    \end{itemize}
The procedure stops after 100 generations or 3 days of computing.
The cross-over rate is set at 0.9 and use 2 points.
The mutation rate is set at 0.1 (a higher mutation did not
implyied a faster convergence nor better results) and apply gaussian mutations.
The number of elites is set at 20.
The fitness value of a chromosome is the $rate$ explained
in~(\ref{eq_fitness}) and the
genetic algorithm wants to reduce this rate.

\section{Measure of Performance}
\subsection{Apparatus}
\label{sec_apparatus}
\begin{figure}[!t]
\centering
\includegraphics[width=3.3cm,height=3.3cm]{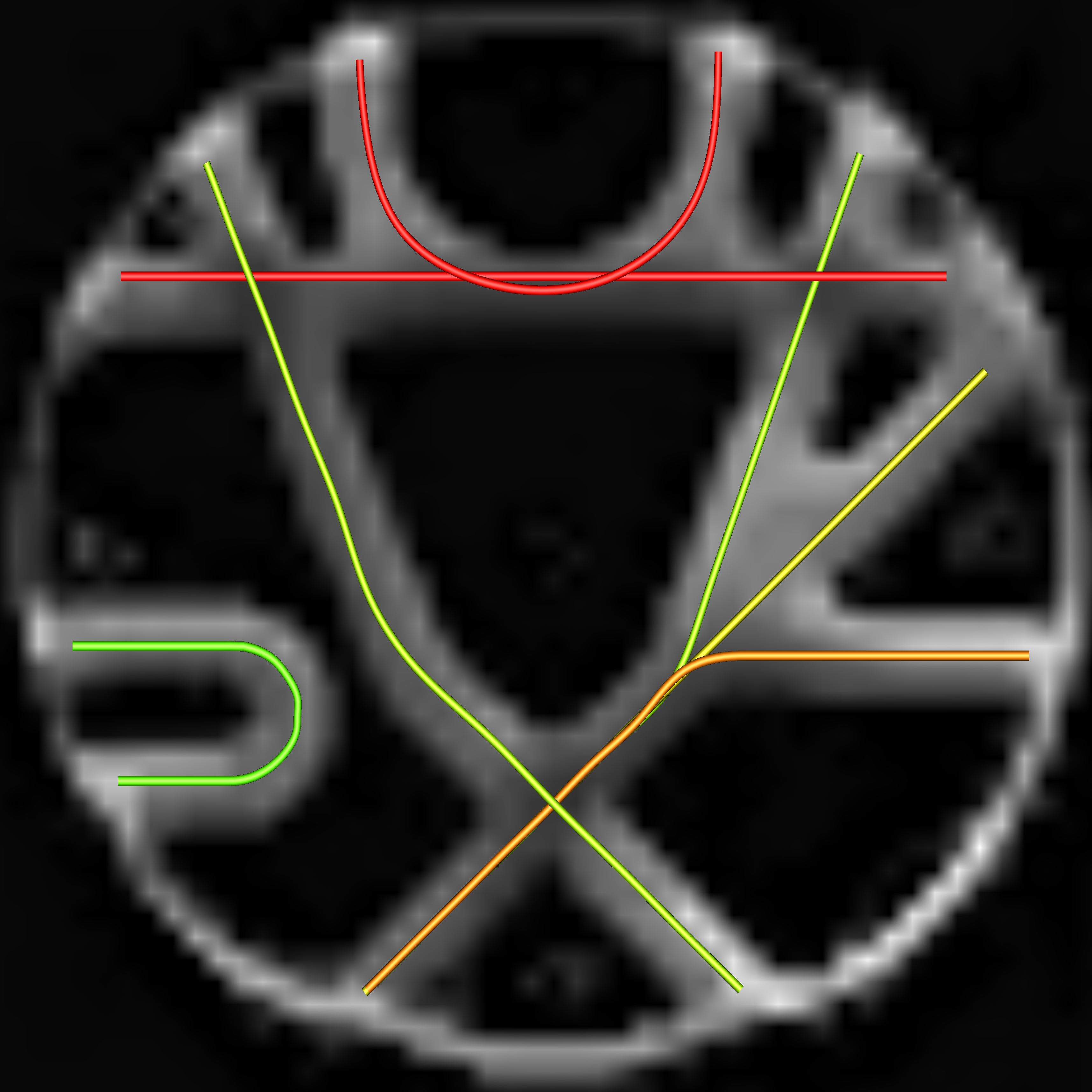} 
\caption{\label{fig:fiberCup}
  Phantom dataset with ground truth fibers.
}
\end{figure}

Ground truth datasets and validation is one of the biggest challenge
of the dMRI community. 
Hence, there is an important 
effort to build  \emph{ex-vivo} phantoms that produce realistic datasets,
more realistic than simulated synthetic data. This is the case of the
FiberCup data, mimicking a coronal slice of the brain. It is a simple
3D dMRI dataset, but is quite unique because it reproduces complex fiber
crossing configurations, similar to configurations in the cemtrum
semioval and 'U' fibers of the brain. The underlying ground truth is
known and thus will serve as our learning/testing dataset. For this
paper, we have focused on the $3$ mm isotropic, 64 directions,
$b=1500$ s/mm$^2$ dataset~\cite{fillard-descoteaux-etal:11}. The
phantom and ground truth fibers are illustrated in
Fig.~\ref{fig:fiberCup}.
Thus, we use a good compromise between synthetic data as
in~\cite{schnell-saur-etal:09} and \emph{in vivo} data for which it
almost impossible to have ground truth. 
Phantom border represents GM, fibers in one direction (resp.
several directions) is WMSF (resp. WMCF), the rest is CSF.

From this HARDI dataset, we provide one input among the following
choices to the SVM: i)-ii) a SH order 4 and 8
representation of raw signal (SH4, SH8
respectively)~\cite{descoteaux-angelino-etal:06c}, or iii) 
the eigenvalues of the diffusion tensor~\cite{Westin2002} (EIG).

\begin{figure*}[tb]
  \centering
  \subfloat[Baseline classifier (SVM fusion of several SVM
  classifiers)]{\label{fig_class_base}\includegraphics[width=.5\linewidth]{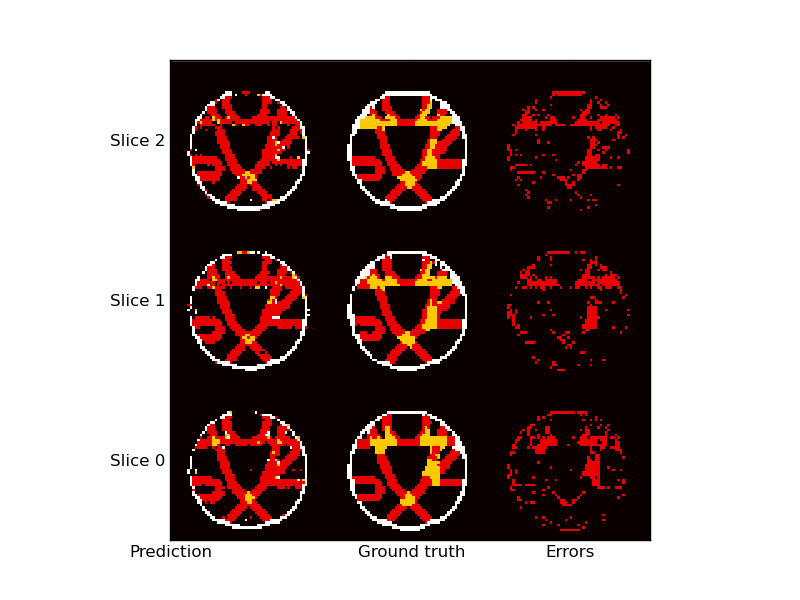}}
%
  \subfloat[Proposal (SVM on convolved
  voxels)]{\includegraphics[width=.5\linewidth]{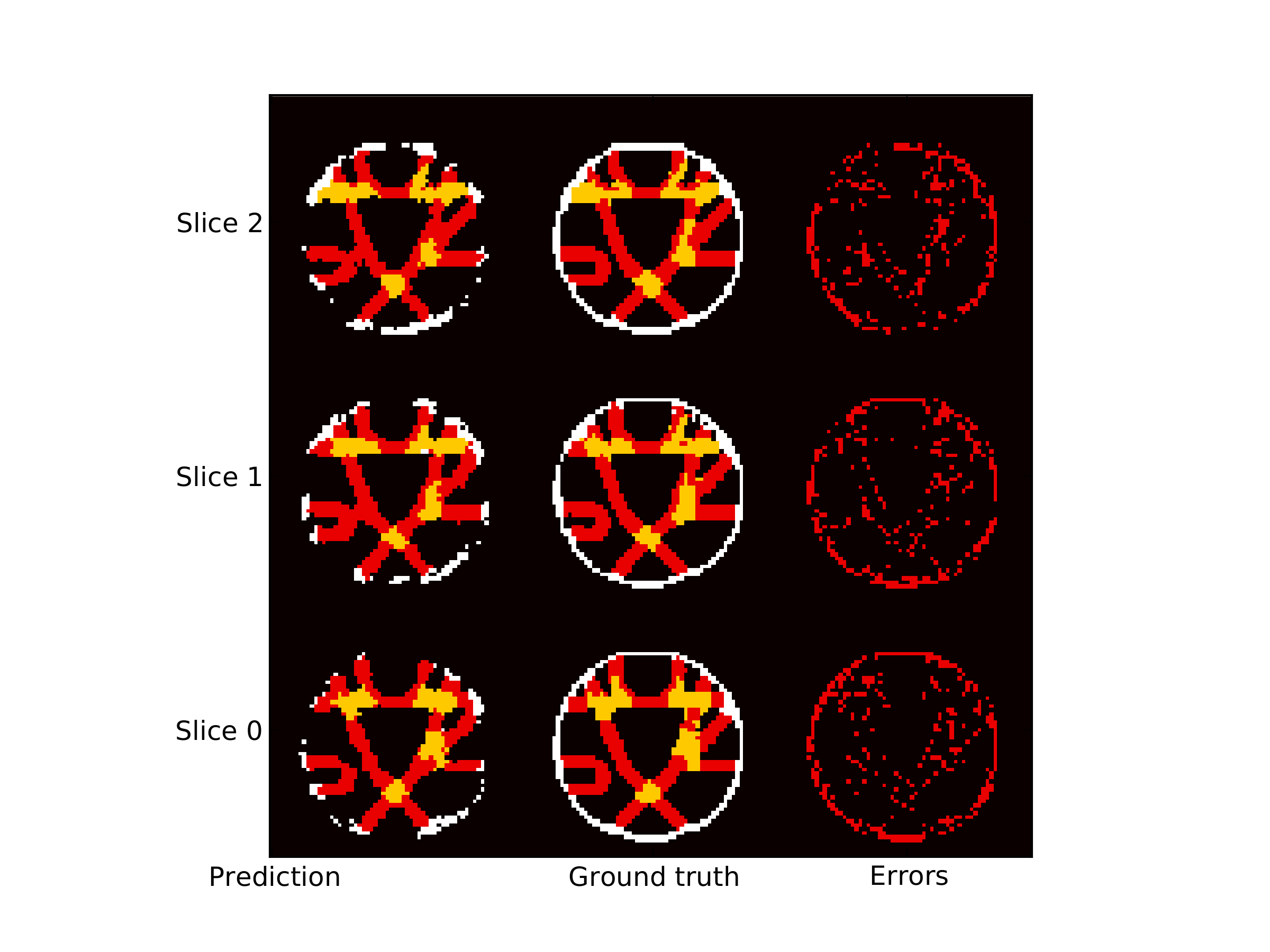}}

  \caption{Visual analysis of the performances of classification.
  For (a) and (b):
  left column is the obtained labels;
  center column is the ground truth;
  right column shows the errors;
  each line is a different slice of the phantom.
  }
  \label{fig_comp_performance}

\end{figure*}

\begin{figure}[tb]
  \centering
\includegraphics[width=.8\linewidth]{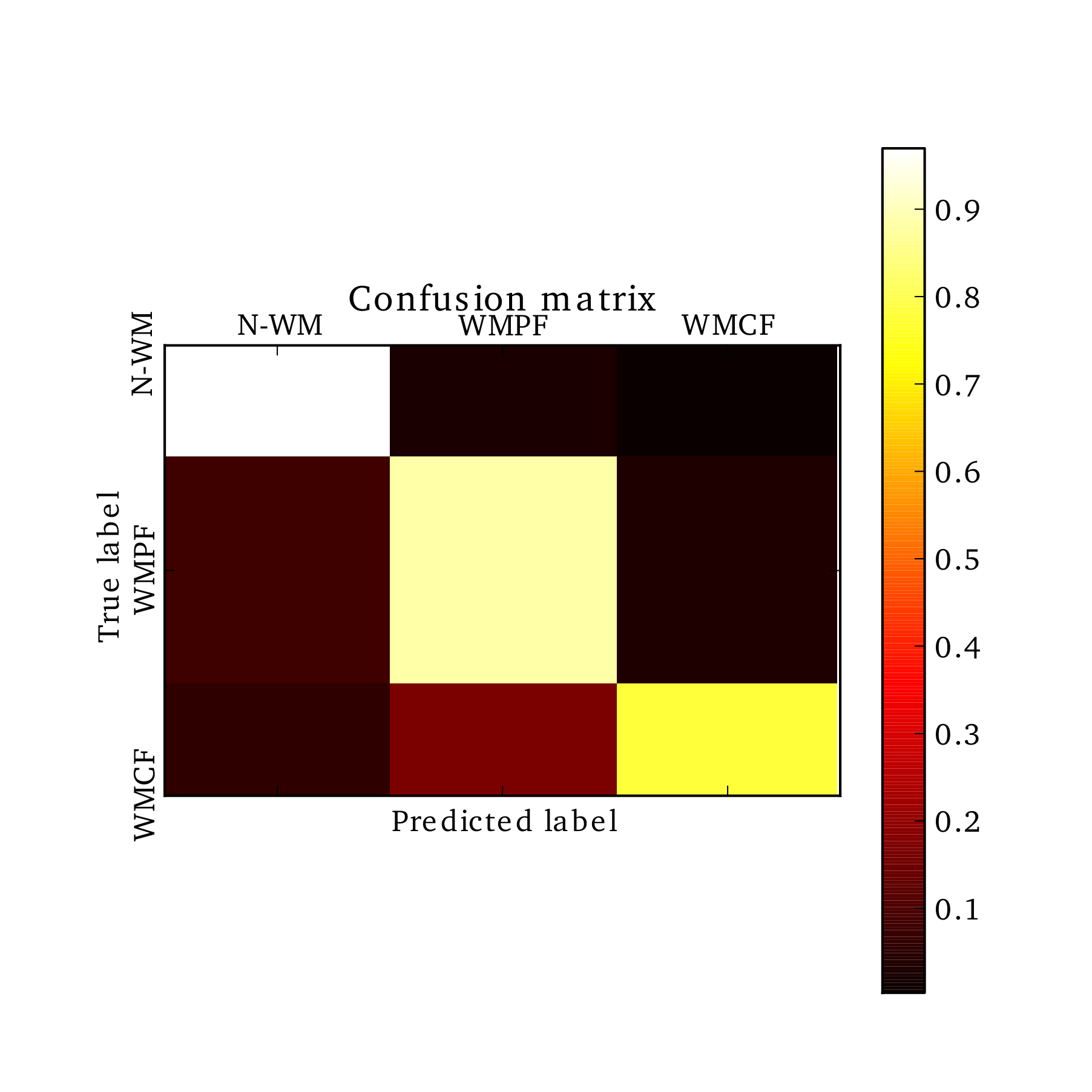}
\caption{Confusion matrix by (merging CSF and GM labels)}
\end{figure}

Results of the proposed method are compared to a baseline classifier.
This classifier does not use convolution operations, but it works
with far more information.
Each voxel is classified by several SVM classifier using a
different feature space (SH4, SH4 rotation
insensitive~\cite{schnell-saur-etal:09}, SH8, SH8
rotation insensitive, eigenvalues SH4 after deconvolution, 
ODF4, ODF8).
Each classifier is trained using the best parameters
($C$,$\gamma$) by getting them through a grid search and 10 fold cross validation.
A final SVM classifier operates a fusion (in order to obtain
better results than a single SVM) of the results of the
previous classifier in order to give the label of the voxel.
Note that this SVM fusion gives better results than a majority
vote and that each individual classifier.

\begin{table}
  \centering
  \caption{Recognition performance (fitness value/global classification
  error rate), 
  using the best convolution
  kernels, and for the baseline classifiers}
  \label{tab_perf_results}
  \begin{tabular}{@{}llll@{}}
        \toprule
	& \multicolumn{2}{c}{ Kernel width}\\
	\cmidrule{2-4}
	Features       & 5                    & 7           & 9 \\
	\midrule
	SH4            & (0.28/0.19)          & (0.29/0.16) & (0.26/0.15) \\
	SH8            & (\textbf{0.21}/0.14) & (0.27/0.17) & (0.23/\textbf{0.12}) \\
	EIG            & (0.42/0.25)          & (0.44/0.8)  & (0.40/0.22)\\
    
    \midrule
    \multicolumn{2}{c}{Baseline majority vote} &
    \multicolumn{2}{c}{Baseline SVM }\\ 

    \multicolumn{2}{c}{(0.42/0.18)} &
    \multicolumn{2}{c}{(0.36/0.16) }\\ 
    \bottomrule
  \end{tabular}
\end{table}

\subsection{Results}
  Using a computer having 4Gb of RAM and a processor of 4 cores,
  scripts written in python and consuming tasks compiled using
  Cython, it takes around 24 hours
  to run the evolution procedure
  for most couples of feature/kernel width (we need to evaluate
  $50000=500*100$ classification tasks on a high quantity of
  voxels for each couple of selected feature and kernel width).
  As the feature space is smaller, results or obtained quickly for
  SH4 data than for SH8.

  The evolution procedure was not able to find a better individual
  than the standard kernel matrices for some configurations.
  This can be because of the size of the search space which is too
  big in comparison to the size of the quantity of individuals
  involved in the procedure.

  Tab.~\ref{tab_perf_results} presents the performances obtained
  using the best filter sets for each couple of features (SH4,
  SH8) and convolution kernel's width (5, 7, 9) as well as the
  performances of the baseline classifiers.
  Fig.~\ref{fig_comp_performance} visually presents the
  recognition performance using our method (SH8, kernel width of
  5) against the best baseline classifier.

  Baseline classifiers mainly do mistakes by detecting crossing
  fibers as single fibers.
  If we use the global classification error rates, they seem to
  perform well, however if we use the fitness value or look at
  Fig.~\ref{fig_class_base}, we understand they perform badly.
  This may be explained because it is trained in order to reduce
  the global error rate whatever is the location of the errors.
  The proposed convolution based classifier performs better.
  Most error are miss recognition of the border of the phantom.
  This can be explained by the fact that the fitness function does
  not try to minimize this error.
  Most of the other errors are located at the frontier between two
  different classes.
  If we ignore the recognition error between
  CSF and GM (because there may not be important in our context), the
  error rate drops from 14.65\% to 6.61\% (SH8/window of size 5).

  The huge amount of time involved in the experiment is not representative of a real
  use of the system as a lot of classifiers are learned and tried
  during the optimisation procedure.
  To give a more accurate representation of the duration of the
  computation,
  we have computed the mean time taken to learn the classifier and
  the mean time taken to classify the voxels by using the best
  kernels for each configuration of features and kernel width.
  It takes in average less than 11s to learn the classifier and
  a bit more than 1.50s to classify each voxel.
  In a real life application,  only the
  classification process is used ; so we think it is fast enough to
  be used in a real world application even if processing time
  would be slightly larger with more voxels.
  We can approximate an overestimate of the classification duration of a whole brain
  by:
  \begin{equation}
    timing(V) = \frac{1.5*V}{3*64^2}
  \end{equation}
  \noindent with V the number of voxels to classify.
  It is an overestimate as it does not take into account a
  potential reduction of the number of voxels to classify after
  having applyied a manual or automatic localisation of the brain.
  Thus with a matrix of $256*256*80$, the classification duration
  would be approximatly $timing(5242880)=640.0s$. Ten minutes
  seems to be a correct amount of time.


\section{Conclusion}
We have proposed a new SVM classifier taking into account a local
spatial neighbourhood to classify each voxel of HARDI data according
to its water diffusion phenomenon. We can successfully classify voxels
containing single
fiber bundles,  voxels with 
several directions of diffusion reflecting crossing fiber populations
and random isotropic voxels without fiber bundles. 

We believe that this opens many possible perspectives in quantitative
white matter analysis in healthy and patients with neurodegenerative
diseases. Future experiment could use large margin
filtering~\cite{ieeesp2012} in order to optimize the SVM parameters
while optimizing the convolution matrices.
It is also important to apply the results on real
brain datasets manually labelled by neurologists or neurosurgeons.
Morphological operators could also decrease the recognition
error rate.

\balance

\end{document}